\documentclass[journal]{IEEEtran}
\usepackage{latexsym}

\usepackage{enumerate}
\usepackage{graphicx}
\usepackage{subfigure}
\usepackage{multirow}

\usepackage{framed,multirow}
\usepackage{times}
\usepackage{graphicx} 
\usepackage{subfigure}
\usepackage{amssymb}
\usepackage{latexsym}

\usepackage{algorithm}
\usepackage{algorithmic}

\usepackage{hyperref}

\usepackage{url}
\usepackage{xcolor}
\usepackage{amsmath}
\usepackage{amsxtra}
\usepackage{braket}
\usepackage{array}
\usepackage{color}
\usepackage{url}
\usepackage{placeins}

\usepackage{times}
\usepackage{balance}
\usepackage{enumerate}
\usepackage{multirow}
\usepackage{pifont}
\usepackage{marvosym}
\usepackage{ifsym}
\usepackage{threeparttable}

\usepackage{lipsum} 
\usepackage{amsmath,esint} 
\usepackage{cuted}

\begin{document}
\title{HC-GAE: The Hierarchical Cluster-based Graph Auto-Encoder for Graph Representation Learning}
\author{Zhuo~Xu,~\IEEEmembership{}
        Lu~Bai${}^{*}$,~\IEEEmembership{IEEE~Member}
        Lixin~Cui,~\IEEEmembership{IEEE~Member}
        Ming~Li,~\IEEEmembership{IEEE~Member}\\
        Yue~Wang,~\IEEEmembership{}
        Edwin R.~Hancock,~\IEEEmembership{IEEE~Fellow}

\thanks{

Zhuo Xu and Lu Bai (${}^{*}$Corresponding Author: bailu@bnu.edu.cn) are with School of Artificial Intelligence, Beijing Normal University, Beijing, China. Lixin Cui and Yue Wang are with School of Information, Central University of Finance and Economics, Beijing, China. Ming Li is with the Key Laboratory of Intelligent Education Technology and Application of Zhejiang Province, Zhejiang Normal University, Jinhua, China. Edwin R. Hancock is with Department of Computer Science, University of York, York, UK}
}
\markboth{}%
{Shell \MakeLowercase{\textit{et al.}}: Bare Demo of IEEEtran.cls for Journals}

\maketitle

\begin{abstract}
Graph Auto-Encoders (GAEs) are powerful tools for graph representation learning. In this paper, we develop a novel Hierarchical Cluster-based GAE (HC-GAE), that can learn effective structural characteristics for graph data analysis. To this end, during the encoding process, we commence by utilizing the hard node assignment to decompose a sample graph into a family of separated subgraphs. We compress each subgraph into a coarsened node, transforming the original graph into a coarsened graph. On the other hand, during the decoding process, we adopt the soft node assignment to reconstruct the original graph structure by expanding the coarsened nodes. By hierarchically performing the above compressing procedure during the decoding process as well as the expanding procedure during the decoding process, the proposed HC-GAE can effectively extract bidirectionally hierarchical structural features of the original sample graph. Furthermore, we re-design the loss function that can integrate the information from either the encoder or the decoder. Since the associated graph convolution operation of the proposed HC-GAE is restricted in each individual separated subgraph and cannot propagate the node information between different subgraphs, the proposed HC-GAE can significantly reduce the over-smoothing problem arising in the classical convolution-based GAEs. The proposed HC-GAE can generate effective representations for either node classification or graph classification, and the experiments demonstrate the effectiveness on real-world datasets.
\end{abstract}

\begin{IEEEkeywords}
Graph Auto-Encoder; Graph Neural Networks; Graph Classification; Node Classification
\end{IEEEkeywords}

\maketitle
\IEEEpeerreviewmaketitle

\section{Introduction}\label{intro}

In real-world applications, graph structure data has been widely used for characterizing pairwise relationships among the components of complex systems. With the recent rapid development of deep learning, the graph representation learning approaches relying on neural networks are introduced for the analysis of various graph data, e.g., social networks~\cite{social_network}, transportation networks~\cite{traffic}, protein compounds~\cite{protein}, etc. One challenging arising in these studies is that the graph data has a nonlinear structure defined in an irregular non-Euclidean space, and it is hard to directly employ traditional neural networks to learn graph representations.

To overcome the above problem, there have been increasing interests to further generalize traditional neural networks, especially the Convolutional Neural Network (CNN)~\cite{CNN}, for the irregular graph data. These are the so-called convolution-based Graph Neural Networks (GNNs)~\cite{GNN} and their related approaches~\cite{GCN} proposed for graph-based tasks, utilizing both graph features and topologies. For instance, the Higher-order Graph Convolutional Network (HiGCN)~\cite{HiGCN} has been developed based on the higher-order interactions to recognize intrinsic features across varying topological scales. Its effective expressiveness makes it capable for various graph-based tasks. The DeepRank-GNN~\cite{bioinfor} has been proposed by combining the rotation-invariant graphs and the GNN to represent protein-protein complexes. Because the GNN models can extract graph representations with more semantic learning under supervised conditions, researchers have focused more on seeking a self-supervised framework associated with the GNNs to accomplish representation learning.

As a typical framework of representation learning, the classical Auto-Encoder~\cite{autoencoder} has been proposed to extract impressive results by reconstructing the input information. Especially, the Graph Auto-Encoder (GAE)~\cite{GAE} associated with the GNN model has further generalized the reconstruction ability for graph structures~\cite{GAE_feature,GAE_connection}. Due to the extensibility, the GAEs have been developed as a family of classical models for self-supervised representation learning, and there are adequate derivation models belonging to GAEs. For instance, the Self-Supervised Masked Graph Autoencoders (GraphMAE)~\cite{GraphMAE} focusing on the feature reconstruction adopts a masking strategy and the scaled cosine error in the training model. Compared to the traditional GAE approach like VGAE~\cite{VGAE}, its decoder is retrofitted with the GNN and the re-masking operation. Based on the GraphMAE, the S2GAE~\cite{S2GAE} continues to adopt the masking strategy to improve the auto-encoder framework. To generate the cross-representation, the decoder is designed to capture the cross-correlation of nodes.

\textbf{Challenges.} Although the classical GAE-based methods achieve the effective performance for graph representation learning, they still have some significant challenging problems summarized as follows.

\textbf{(a) The limitation for multiple downstream tasks}: Generally, the representations extracted from the GAEs can be divided into several categories, including the node-level representations for node classification, the graph-level representations for graph classification, etc. Specifically, it is difficult for the GAEs to generate universal representations for multiple downstream tasks simultaneously. This is because the GAEs tend to over-emphasize the node features. For instance, the GraphMAE~\cite{GraphMAE} focuses more on the node feature reconstruction, resulting in \textbf{topological missing} and weakening the the structure information reconstruction. This is harmful for the graph-level representation learning.

\textbf{(b) The over-smoothing problem}: The GAEs are usually proposed based on the GNNS, thus both the decoder and encoder modules of the GAEs are defined associated with a number of stacked graph convolution operations, that rely on the node information propagation between adjacent nodes. When the GAE becomes deeper, the node features tend to be similar or indistinguishable after multiple rounds of information passing~\cite{Rethinking}, resulting in the notorious over-smoothing problem~\cite{deep_over_smoothing} and influence the performance of the GAEs.



\textbf{Contributions.} The aim of this paper is to overcome the above challenging problems by proposing a novel HC-GAE model. Overall, the main contributions are threefold.

\textbf{First}, we propose a novel Hierarchical Cluster-based GAE (\textbf{HC-GAE}) for graph representation learning. Specifically, for the encoding process, we adopt the hard node assignment to decompose a sample graph into a family of separated subgraphs. We perform the graph convolution operation for each subgraph to further extract node features and compress the nodes belonging to each subgraph into a coarsened node, transforming the original graph into a coarsened graph. Since the separated subgraphs are isolated from each other, the convolution operation cannot propagate the node information between different subgraphs. The proposed HC-GAE can in turn reduce the over-smoothing problem arising in the classical GAEs. Moreover, since the effect of the graph structure perturbation is limited within each subgraph, the required convolution operation performed on each subgraph can strengthen the robustness of the encoder for the proposed HC-GAE. As a result, the outputs of the encoder can be employed as the graph-level representations. On the other hand, for the decoding process, we adopt the soft node assignment to reconstruct the original graph structure by expanding each coarsened node into all retrieved nodes probabilistically. Thus, the outputs of the decoder can be employed as the node-level representations. Since the HC-GAE is defined by hierarchically performing the above compressing procedure during the decoding process as well as the expanding procedure during the decoding process, the proposed HC-GAE can effectively extract bidirectionally hierarchical structural features of the original sample graph, resulting in effective hierarchical graph-level and node-level representations for either graph classification or node classification.

\textbf{Second}, we propose a new loss function for training the proposed HC-GAE model. For calculating the complete loss value, we integrate the local loss from the subgraphs in the encoding operation and the global loss from the reconstructed graphs in the decoding operation. The global loss can capture the information from both the structure and the feature reconstruction processes. The combination of these two pretext tasks broadens the strict requirement causing the topological closeness. In addition, to avoid the over-fitting problem, we add the local loss as the regularization in our loss function.

\textbf{Third}, we empirically evaluate the performance of the proposed HC-GAE model on both node and graph classification tasks, demonstrating the effectiveness of the proposed model.

\section{Related Works}\label{related_work}

\subsection{Graph Neural Network}\label{GNN}

GNNs are widely utilized across adequate application scenarios~\cite{gnn_link_pre,gnn_image_cls,gnn_graph_cls}, and achieves a prominent success. The input data of GNNs is graphs, a kind of non-Euclidean data, containing nodes and edges. With the complex structure of the graphs, GNNs aim to leverage the information passing mechanism among nodes for graph embedding learning. The process of information passing could be divided into aggregating, combining and readout.

Given an input graph $G(V,E)$ with the node set $V$ and the edge set $E$, the node information is represented as the feature matrix $X \in \mathbb{R}^{n \times d}$ with $d$ features, and the structure information is represented as the adjacent matrix $A \in \{0,1\}^{n \times n}$. The GNN for each layer is defined as
\begin{equation}
    Z_{G} = \mathrm{GNN} (X,A;\Theta),
\end{equation}
where $Z_{G}$ is the graph embedding, and $\Theta$ is the parameter set of the GNN. This embedding result is used for the downstream tasks. Its methodology researches could be categorized into the spectral and spatial approaches~\cite{gnn_survey}. When computing power is not enough to realize operations on graph, there are several researches focusing on the graph spectral domain~\cite{spectral_dev}.


Graph convolutional networks (GCNs)~\cite{GCN}, a typical derivative model of GNNs, generalize convolutional neural networks (CNNs)~\cite{CNN} to the graph-structured data. They have performed in various graph application tasks~\cite{gcn_apply}.And they are widely utilized in deep learning models~\cite{hierarchical_gcn_1,hierarchical_gcn_2}. For example, the GCN proposed by Kipf et al.~\cite{GCN}, adopts the following layer-wise scheme to realize the hierarchical model, i.e.,
 \begin{equation}
    H^{(l+1)} =  \mathrm{ReLU}(\tilde{D}^{- \frac{1}{2}}\tilde{A}\tilde{D}^{- \frac{1}{2}}H^{(l)}W^{(l)}),\label{GCN_module}
\end{equation}
where $H^{(l)} \in \mathbb{R}^{n \times d}$ is the hidden embedding matrix in the $l$ layer. $W^{l} \in \mathbb{R}^{d \times d}$ is the trainable matrix in the $l$ layer, $\tilde{A} = A + I$ is the adjacency matrix associated with the self loop, the degree matrix $\tilde{D} = \sum_{j}\tilde{A}_{ij}$ is the corresponding degree matrix, and $H^{(l+1)}$ is the embedding matrix extracted for the next layer $l+1$ of the hierarchical model. Compared to the traditional GNNs, the hierarchical GCN could capture the global representations through multi-layer passing. However, as the hierarchical GCN deepens, the node information is propagated to the whole graph. The over-smoothing problem where the node representations of the graph tend to be similar is obvious in the multi-layer GCN.

\subsection{Graph Auto-Encoder}
\label{GAE}

The GAE is a classical self-supervised framework that completes graph representation learning task. The earliest works related to GAEs are DeepWalk~\cite{deepwalk} and Node2Vec~\cite{node2vec} where encoders play an important role in learning latent representations of vertices. With the addition of GNNs, encoders in GAEs have the ability to cope with non-Euclidean data~\cite{reconstruction}. As a self-supervised learning model, the pretext task of GAEs in training is the graph reconstruction~\cite{little_exp}. In detail, the  reconstruction targets could be categorized into fine-grained and coarse-grained ones.

The fine-grained targets contain either nodes or edges. For instance, the Variational Graph Auto-Encoder (VGAE) model~\cite{VGAE} adopts two stages including encoder and decoder to accomplish representation learning. Assume an input graph $G = (V,E)$, the goal of the VGAE is to embed the graph following the encoder function $f : V \times E \rightarrow Z \in \mathbb{R}^{n \times d}$ which is the mapping from the node set $V$ which has $n$ nodes with $d$ features to the embedding matrix $Z$. Then, the decoder reconstructs the graph through the network $g : Z \rightarrow E^{\prime}$ where $E^{\prime}$ is the reconstructed edge set. The training process of the VGAE is as
\begin{equation}
\label{standard_GAE}
    Z = f(V,E), E^{\prime} = g(Z).
\end{equation}

In the encoding and decoding processes, the VGAE obtains the conditional probability $q(Z \mid V, E)$ from the encoder and $p(E^{\prime} \mid Z)$ from the decoder. The loss function is defined as
\begin{equation}
    \label{equa:VGAE_loss}
    \mathcal{L} = \mathrm{KL}[q(Z \mid V, E)\|p(Z)]-\mathbb{E}_{q(Z \mid V, E)}[\mathrm{log} p(E^{\prime} \mid Z)],
\end{equation}
where $\mathrm{KL}[\cdot]$ is the Kullback-Leibler divergence, $\mathbb{E}$ is the expaction, and $p(Z)$ is the Gaussian prior.

Self-Supervised Graph Autoencoder (S2GAE) proposed by Tan et al.~\cite{S2GAE}, randomly mask a portion of edges and then learn to reconstruct the missing edges. Self-supervised Masked Graph Autoencoders (GraphMAE)~\cite{GraphMAE} also utilizes the masking strategy to reconstruct node features. These methods focus on the local information and disregards several challenges such as over-smoothing.

The coarse-grained targets contain the whole graph, the subgraphs or the paths of the graph. For example, Heterogeneous Graph Masked Autoencoder (HGMA)~\cite{GAE} adopts the dynamic masking strategy to mask the nodes, and edges in the paths and then complete path reconstruction. MaskGAE~\cite{Maskgae} aims to reconstruct the masked edges and node degrees jointly. Recently, some researchers have noted that the combination of Graph Contrastive Learning (GCL)~\cite{gcl} and the GAE framework could realize the capture of complex interdependency in graphs. Self-supervised Learning for Graph Anomaly Detection (SL-GAD)~\cite{SL_GAD} obtains double subgraphs through the graph view sampling, and then respectively reconstructs them in two decoders for constrastive learning.

Although all the methods realize the improvement of representation learning, the expressiveness of the representations is still weak. The aforementioned GAEs rarely notice their limitations in learning schemes.
For the specific downstream task such as node classification, the GAEs could have a nice performance due to the high focus on the node feature reconstruction. This phenomenon where models focus on the node features is named as the topological missing. Since the graph features are over-emphasized, the GAEs are weak in graph structure reconstruction. And these models could be limited in multiple downstream tasks~\cite{little_exp}. Meanwhile, the problem of GNNs mentioned in Section~\ref{GNN} affects the feature learning in GAEs. Especially, over-smoothing caused by information passing affects the GNN encoder. When the perturbation of the graph structure is conducted, the noise could be propagated to the neighbor nodes through the edges. After several rounds of information passing, the generated graph representations are noisy.

\textbf{Current Challenges.} The graph representation learning based on the GAE framework has achieved a nice performance. However, the researchers are disturbed by two problems including \textbf{(a)} \textbf{limitation for multiple downstream tasks}, \textbf{(b)} \textbf{over-smoothing}. These problems have widely existed in the current GAEs. Note that, some models might overcome one of these challenges, but \textbf{cannot} solve them \textbf{simultaneously}.

\section{The Methodology}
\label{methodology}
\begin{figure*}
\vspace{-0pt}
    \centering
    \includegraphics[width=0.85\textwidth]{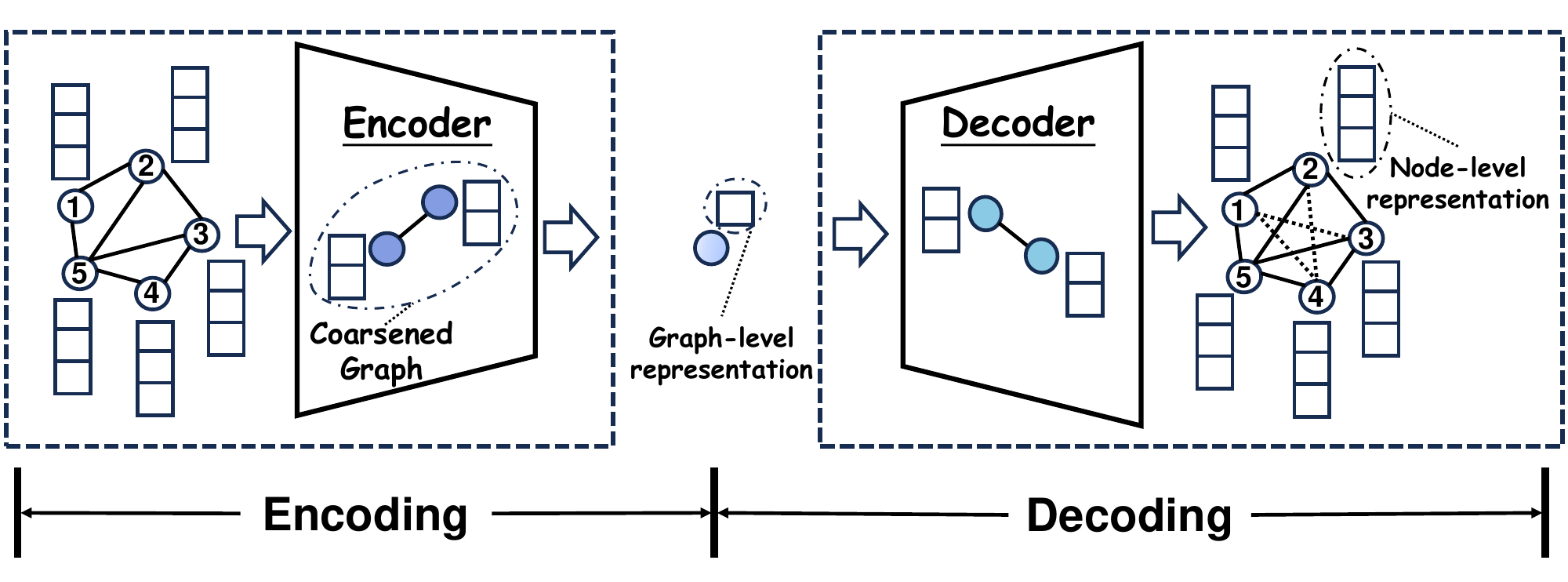}
    \vspace{-10pt}
    \caption{The architecture of our proposed model, HC-GAE.}
    \label{fig:model}
    \vspace{-0pt}
\end{figure*}

To overcome the aforementioned challenges, we propose a novel (\textbf{HC-GAE}) to learn effective graph representations. The overview of our model is shown in Figure~\ref{fig:model}. Similar to the other GAEs, our model has two stages including encoder and decoder. In the encoder, the input graphs are compressed into coarsened graphs through multi-layers.  The results of encoder are the graph-level representations for the graph classification. Then, in the decoding process, the decoder reconstructs the graphs, and outputs the node-level representations for the node classification.

In the following subsections, we first give our proposed GNN encoder and introduce the subgraphs utilized in the encoder. Then, we introduce the GNN decoder with the soft assignment. Compared to the standard GAE loss, our proposed loss calculation is proposed for effective training. At last, we discuss the theoretical properties of our proposed HC-GAE.

\subsection{The GNN Encoder with the Separated Subgraphs}
\begin{figure*}
\vspace{-0pt}
    \centering
    \includegraphics[width=.5\textwidth]{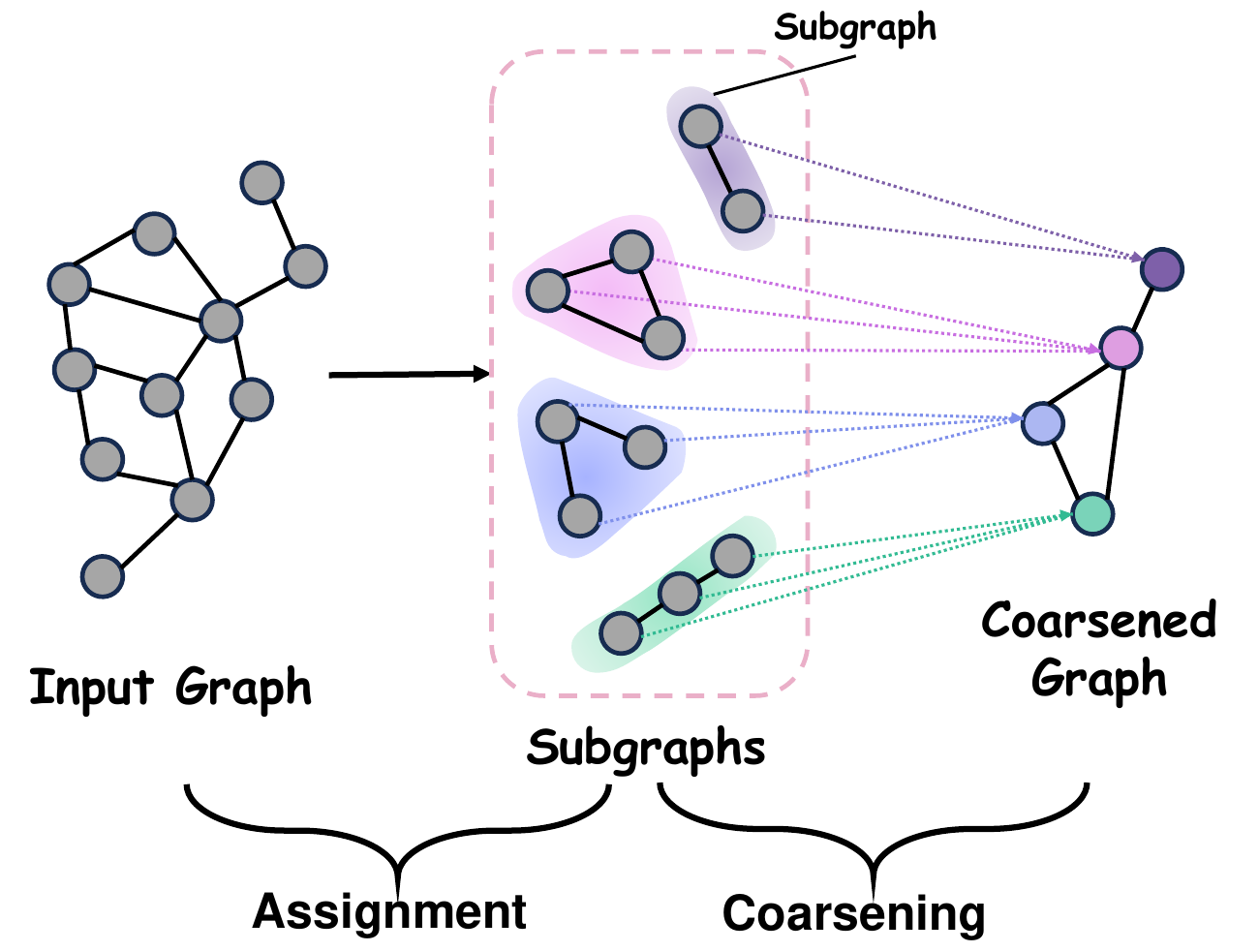}
    \vspace{-10pt}
    \caption{The computational architecture for our proposed layer in the GNN encoder.}
    \label{fig:encoder_layer}
    \vspace{-0pt}
\end{figure*}

The first module of our model is the GNN encoder, which adopts the hierarchical architecture to compress the input graph. The GNN encoder is composed of multiple layers which continuously compress the features and the nodes in graph. The details of our proposed layer in the GNN encoder are shown in Figure~\ref{fig:encoder_layer}. Each layer could be divided into two processes including assignment and coarsening. The first one is to generate subgraphs from the original graph. And these subgraphs map to the nodes of the coarsened graph.

\textbf{Assignment.} For each layer $l$ of encoder, an input graph is denoted as $G^{(l)} = (X^{(l)}, A^{(l)})$ where $X^{(l)} \in \mathbb{R}^{n_{(l)} \times d_{(l)}}$ is the feature matrix and $A^{(l)} \in \mathbb{R}^{n_{(l)} \times n_{(l)}}$ is the adjacent matrix. The number of nodes in $G^{(l)}$ is $n_{(l)}$, and each node has $d_{(l)}$ features. Note that, $G^{(l)}$ could be the original input graph when $l = 1$ or the coarsened graph when $l > 1$. In the assignment process, the graph $G^{(l)}$ is decomposed into subgraphs. And we realize the node assignment through hard assignment where each node cannot be assigned to multiple subgraphs. Given the feature matrix $X^{(l)}$ and the adjacent matrix $A^{(l)}$, we first calculate the soft assignment matrix $S_{\mathrm{soft}}$ which allows a node to assign various subgraphs as follows,

\begin{equation}
    \label{eqa:soft_assign}
    S_{\mathrm{soft}} =
	\begin{cases}
		\mathrm{softmax}(\mathrm{GNN}(X^{(l)}, A^{(l)}))   \ & \mathrm{if} \ l = 1\\
		\mathrm{softmax}(X^{(l)})   \ & \mathrm{if} \ l > 1
	\end{cases},
\end{equation}

where $S_{\mathrm{soft}} \in \mathbb{R}^{n_{(l)} \times n_{(l+1)} }$ and $n_{(l+1)}< n_{(l)}$. Based on the $S_{\mathrm{soft}}$, the $(i,j)$-th entry of the hard assignment matrix $S^{(l)}\in \{0, 1\}^{n_{(l)} \times n_{(l+1)}}$ satisfies

\begin{equation}
\label{equa:assignment_equation}
     S^{(l)}(i,j) =
     \begin{cases}
		1   \ & \mathrm{if} \ S_{\mathrm{soft}}(i,j) = \mathop{\max}\limits_{\forall j \in n_{l+1}}[S_{\mathrm{soft}}(i,:)]\\
		0   \ & \mathrm{otherwise}
	\end{cases}.
\end{equation}

Clearly, each $i$-th row of the hard assignment matrix $S^{(l)}$ selects the maximum element as $1$ and the remaining elements as $0$, i.e., the $i$-th node is only assigned to the $j$-th subgraph. And the $j$-th subgraph is denoted as $G_{j}^{(l)}(V_{j}^{(l)},E_{j}^{(l)})$ where $V_{j}^{(l)}$ is the node set including the nodes of $G_{j}^{(l)}$ and $E_{j}^{(l)}$ is the edge connections of nodes in $V_{j}^{(l)}$.

\textbf{Coarsening.} Based on the generation of the subgraph, the coarsening process aims at compressing these subgraphs into nodes in the coarsened graph. Given the associated feature matrix $X_{j}^{(l)}$ and adjacent matrix $A_{j}^{(l)}$ of the subgraph $G_{j}^{(l)}$, we adopt a local graph coarsening operation to extract the local information as

\begin{equation}
    Z^{(l)}_{j} = A_{j}^{(l)}X_{j}^{(l)}W_{j}^{(l)},
\end{equation}
where $W_{j}^{(l)} \in \mathbb{R}^{d_{(l)} \times d_{(l+1)}}$ ($d_{(l)} > d_{(l+1)}$) is the trainable weight matrix of layer $l$, and $Z^{(l)}_{j} \in \mathbb{R}^{|V_{j}^{(l)}| \times d_{(l+1)}}$ represents the resulting matrix of subgraph $G_{j}^{(l)}$. To compress each subgraph  $G_{j}^{(l)}$, we utilize a mapping vector $\textbf{s}^{l}_{j}$, and
\begin{equation}
\label{equa:mapping_vec}
    \textbf{s}^{(l)}_{j} = \mathrm{softmax}(A_{j}^{(l)}X_{j}^{(l)}D_{j}^{(l)}),
\end{equation}
where $D_{j}^{(l)} \in \mathbb{R}^{d_{(l)} \times 1}$ is the training vector. $\textbf{s}^{l}_{j}$ plays an important role in compressing each $j^{\mathrm{th}}$ subgraph $G_{j}^{(l)}$ to the node of the coarsened graph. After several local graph operations on the separated subgraphs in the $l$-th layer, we aggregate these local information to further generate the coarsened graph, as the input graph $G^{(l+1)}$ for the next layer. We collect the feature matrices of the $l$-th subgraphs as the feature matrix $Z^{(l)} \in \mathbb{R}^{n_{(l)} \times d_{(l+1)}}$ whose node sequence follows the original input graph $G^{(l)}$. In detail, each vertex feature vector of $Z^{(l)}$ is equal to that of the corresponding node in $G_{j}^{(l)}$, since each original node in $G^{(l)}$ essentially corresponds the embedding node of subgraph $G_{j}^{(l)}$. Given the hard assignment matrix $S^{(l)}$ defined by Equation~\ref{equa:assignment_equation}, the mapping vector $\textbf{s}^{(l)}_{j}$ of each subgraph $G_{j}^{(l)}$ defined by Equation~\ref{equa:mapping_vec}, the feature matrix $Z^{(l)}$ and the adjacent matrix $A^{(l)}$, we calculate the feature matrix $X^{(l+1)}$ and the adjacent matrix $A^{(l+1)}$ of the resulting coarsened graph $G^{(l+1)}$ as
\begin{equation}
        X^{(l+1)} = \mathrm{Reorder} [\mathop{\|} \limits_{j=1}^{n_{l+1}} \textbf{s}^{(l)^{\top}}_{j}] Z^{(l)},
\end{equation}
and
\begin{equation}
    A^{(l+1)} = S^{(l)^{\top}}A^{(l)}S^{(l)},\label{A_equation}
\end{equation}
where $\mathop{\|}$ is a concatenation operation, and the function $\mathrm{Reorder}$ reorders the sequences of $[\cdot]$ to follow the same node order of $G_{j}^{(l)}$. In our model, we set the number of the encoder layer to $L$. And the resulting graph $G^{(L+1)} = (X^{(L+1)}, A^{(L+1)})$ is the input graph for decoder. Meanwhile, we adopt a non-parameterized readout function (e.g., MaxPooling and MeanPooling) to embed $X^{(L+1)}$ into the graph-level representation for graph classification.

\subsection{GNN Decoder with the Soft Assignment}
\begin{figure*}
\vspace{-0pt}
    \centering
    \includegraphics[width=0.5\textwidth]{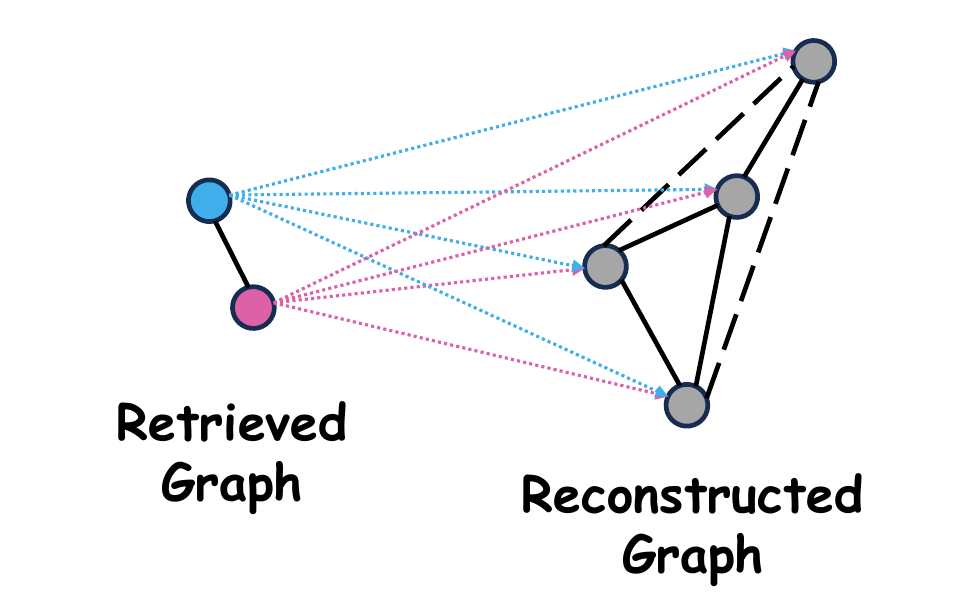}
    \vspace{-10pt}
    \caption{The illustration of our proposed layer in the GNN decoder.}
    \label{fig:decoder_layer}
    \vspace{-0pt}
\end{figure*}

After encoding the input graph, we adopt our proposed GNN decoder which aims at generating the reconstructed graph. Similar to the encoder, our decoder is composed of multiple layers. The details of our proposed layer in the GNN decoder are shown in Figure~\ref{fig:decoder_layer}. The input of the decoder layer is named as the retrieved graph, and the result is defined as the reconstructed graph.  The generation method of the reconstructed graphs is the soft assignment where each input node in the retrieved graph is assigned to the whole reconstructed graph. Then, we introduce how the reconstructed graph is generated from the the retrieved graph.

\textbf{Reconstruction.} We denote the retrieved graph which is the input to $l^{\prime}$-th layer of decoder as $G^{\prime (l^{\prime})} = (X^{\prime (l^{\prime})}, A^{\prime (l^{\prime})})$ where $X^{\prime (l^{\prime})} \in \mathbb{R}^{n_{(l^{\prime})} \times d_{(l^{\prime})}}$ is the feature matrix and $A^{\prime (l^{\prime})} \in \mathbb{R}^{n_{(l^{\prime})} \times n_{(l^{\prime})}}$ is the adjacent matrix. Note that, $G^{\prime (l^{\prime})}$ is equal to $G^{(L)}$ when $l^{\prime} = 1$. Given the input $G^{\prime (l^{\prime})}$ with the feature matrix $X^{\prime (l^{\prime})}$ and the adjacent matrix $A^{\prime (l^{\prime})}$, we denote the learned re-assignment matrix as $\bar{S}^{(l^{\prime})} \in \mathbb{R}^{n_{(l^{\prime})} \times n_{(l^{\prime}+1)}}$ and the embedding matrix as $\bar{Z}^{(l^{\prime})} \in \mathbb{R}^{n_{(l^{\prime})} \times d_{(l^{\prime}+1)}}$ which both are located at the layer $l^{\prime}$. Compared to the complex encoding process, these matrices are calculated as
\begin{equation}
    \bar{S}^{(l^{\prime})} =  \mathrm{softmax}(\mathrm{GNN}_{l^{\prime},\mathrm{re}}(X^{\prime (l^{\prime})}, A^{\prime (l^{\prime})})),
\end{equation}
and
\begin{equation}
    \bar{Z}^{(l^{\prime})} = \mathrm{GNN}_{l^{\prime},\mathrm{emb}}(X^{\prime (l^{\prime})}, A^{\prime (l^{\prime})}),
\end{equation}
where $\mathrm{GNN}_{l^{\prime},\mathrm{re}}$ and $\mathrm{GNN}_{l^{\prime},\mathrm{emb}}$ are two different GNN blocks which do not share parameters. Clearly, although both the GNN blocks have the same inputs, there is an obvious distinction in their functions. In detail, the $\mathrm{GNN}_{l^{\prime},\mathrm{re}}$ generates a probabilistic distribution assigning nodes to the reconstructed graph, while the $\mathrm{GNN}_{l^{\prime},\mathrm{emb}}$ is to generate the new embeddings. With the re-assignment matrix $ \bar{S}^{(l^{\prime})}$ and the embedding matrix  $\bar{Z}^{(l^{\prime})}$,  we calculate the resulting $X^{\prime (l^{\prime}+1)}$ and $A^{\prime (l^{\prime}+1)}$ as
\begin{equation}
    X^{\prime (l^{\prime}+1)} = \bar{S}^{(l^{\prime})^{\top}}\bar{Z}^{(l^{\prime})},
\end{equation}
and
\begin{equation}
    A^{\prime (l^{\prime}+1)} =  \bar{S}^{(l^{\prime})^{\top}}A^{\prime (l^{\prime})}\bar{S}^{(l^{\prime})},
\end{equation}
where $X^{\prime (l^{\prime}+1)} \in \mathbb{R}^{n_{(l^{\prime}+1)} \times d_{(l^{\prime}+1)}} $ is the feature matrix and $A^{\prime (l^{\prime}+1)} \in \mathbb{R}^{n_{(l^{\prime}+1)} \times n_{(l^{\prime}+1)}} $ is the adjacent matrix belonging to the reconstructed graph $G^{\prime (l^{\prime}+1)}$. In our model, we set the number of the decoder layer to $L^{\prime}$, so the resulting graph $G^{\prime (L^{\prime}+1)} = (X^{\prime (L^{\prime}+1)},A^{\prime (L^{\prime}+1)})$ is also the result of our GAE. Note that, the graph $X^{\prime (L^{\prime}+1)}$ is the node-level representations for the node classification task.

\subsection{Our Loss Function}

Compared to the standard loss $\mathcal{L}$ in the Equation~\ref{equa:VGAE_loss}, our loss function improves the calculation to strengthen the expressiveness of representations. In detail, our model is composed of the encoder focusing the local information of subgraphs and the decoder generating the reconstructed graphs. Therefore, our loss could be divided into two parts including the local loss and the global one, i.e.,
\begin{equation}
\begin{aligned}
    \label{equa:our_loss}
    & \mathcal{L}_{\mathrm{local}} = \sum_{l = 1}^{L}\sum_{j=1}^{n_{(l+1)}} \mathrm{KL}[q(Z_{j}^{(l)} \mid X_{j}^{(l)},A_{j}^{(l)}) \| p(Z^{(l)})], \\
    & \mathcal{L}_{\mathrm{global}} = - \sum_{l = 1}^{L}\mathbb{E}_{q(X^{(L)}, A^{(L)}) \mid X^{(l)}, A^{(l)})} \\ 
    &[\mathrm{log}p(X^{\prime(L-l+2)},A^{\prime(L-l+2)} \mid X^{(L)}, A^{(L)}))], \\
    & \mathcal{L}_{\mathrm{HC-GAE}} = \mathcal{L}_{\mathrm{local}} + \mathcal{L}_{\mathrm{global}},
\end{aligned}
\end{equation}
where $\mathcal{L}_{\mathrm{HC-GAE}}$ is our proposed loss, $\mathcal{L}_{\mathrm{local}}$ is the local loss, $\mathcal{L}_{\mathrm{global}}$ is the global loss, and $p(Z^{(l)})$ is the Gaussian prior for the $l$-th layer subgraphs. Compared to the loss $\mathcal{L}$ in the Equation~\ref{equa:VGAE_loss}, our local loss $\mathcal{L}_{\mathrm{local}}$ aims at training the subgraph generation which reserves the local information and avoids over-smoothing in the GNN encoder. And we set the global loss $\mathcal{L}_{\mathrm{global}}$ to train the reconstruction of the graph features and structures. The combination of the local loss $\mathcal{L}_{\mathrm{local}}$ and the global loss $\mathcal{L}_{\mathrm{global}}$ broadens the reconstruction requirement for multiple downstream tasks, since $\mathcal{L}_{\mathrm{local}}$ is a regularization for the loss. Therefore, $\mathcal{L}_{\mathrm{HC-GAE}}$ not only strengthens the graph representations with the additions of the local information, but also addresses the challenges mentioned in Section~\ref{GAE}.

\begin{table*}

\centering
\footnotesize
\caption{Datasets for node classification}\label{table:node_cls_data}
\begin{tabular}{cccccc}
\hline
\textbf{Datasets} & Cora & CiteSeer & ~PubMed & Computers & CS    \\ \hline
\textbf{Nodes}    & 2708 & 3312     & 19717   & 13752     & 18333 \\
\textbf{Edges}    & 5429 & 4660     & 44338   & 245861    & 81894 \\
\textbf{Features} & 1433 & 3703     & 500     & 767       & 6805  \\
\textbf{Classes}  & 7    & 6        & 3       & 10        & 15  \\ \hline
\end{tabular}
\end{table*}

\begin{table*}
\centering
\footnotesize
\caption{Datasets for graph classification}\label{table:graph_cls_data}
\begin{tabular}{cccccc}
\hline
\textbf{Datasets}    & IMDB-B & IMDB-M & PROTEINS & COLLAB  & MUTAG \\ \hline
\textbf{Graphs}      & 1000   & 1500   & 1113     & 5000    & 188   \\
\textbf{Nodes(mean)} & 19.77  & 13     & 39.06    & 74.49   & 17.93 \\
\textbf{Edges(mean)} & 96.53  & 65.94  & 72.82    & 2457.78 & 19.79 \\
\textbf{Classes}     & 2      & 3      & 2        & 3       & 1  \\ \hline
\end{tabular}
\end{table*}
\subsection{Discussions}

\textbf{(a) Why are our results effective in multiple downstream tasks?}

As we mentioned in Section~\ref{intro} and Section~\ref{GAE},  the traditional GAE methods might over-emphasize the goals of the graph reconstruction. The typical result of this operation is topological missing, which damages the graph structure learning and aggravates the backward of the over-fitting in graph features~\cite{drawback_edge}. Therefore, its resulting graph-level representations could not be effective in the graph classification. To generate generalized representations for multiple downstream tasks, our approach proposes some novel operations. And we summarize the reasons why they are effective as follows.

\textbf{First}, our model adopts a series of the assignment strategies to improve the encoding and decoding. In the encoder, we utilize the separated subgraphs to decompose the input graph, and assign the nodes by the hard assignment. In the decoder, the generation of the reconstructed graph follows the soft assignment. The hard assignment reserves the local heterogeneous information and abandons redundancy in the graph-level representation learning. And the soft assignment accomplishes the generation of the node-level representations. The combination of these assignment strategies ensures that our proposed model have the generalized capability to learn multi-level representations for various downstream tasks.  \textbf{Second}, our model re-design the loss function suitable for the training of the new modules.
The global loss $\mathcal{L}_{\mathrm{global}}$ set two reconstruction goals including the graph features and the structure. Multiple goals in self-supervision reduce the over-emphasizing on graph features, that causes the topological missing.
The local loss  $\mathcal{L}_{\mathrm{local}}$ not only plays a role as regularization in $\mathcal{L}_{\mathrm{HC-GAE}}$, but captures the local information from the subgraphs for training. The addition of the local information is a common method to improve the generalization of the graph representations~\cite{ssl_graph}.

\textbf{(b) Why does the over-smoothing hardly affect the model?}

In the encoding process, we adopt separated subgraphs where there is no connection between these structures. Unlike the hierarchical GNN methods such as DiffPool~\cite{DiffPool} or GAE methods such as VGAE~\cite{VGAE}, the message passing is limited in the subgraphs. The node information hardly propagates to the whole graph.  This operation can significantly reduce the over-smoothing problem.

\section{Experiments}
\label{experiments}

In this section, we evaluate the performance of our proposed model over two important graph learning tasks including node classification and graph classification. The details of the datasets and the experiment settings are shown in Table~\ref{table:node_cls_data} and Table~\ref{table:graph_cls_data}.

\begin{table*}
\vspace{-0pt}
\centering
 \footnotesize
\caption{Node classification performance based on accuracy. A.R. is the average rank.}\label{node_cls_res}
\begin{tabular}{ccccccc}
\hline
Datasets          & Cora                  & CiteSeer              & ~PubMed               & Computers             & CS                    & A.R.         \\ \hline
DGI             & 85.41$\pm$0.34        & 74.51$\pm$0.51       & 85.95$\pm$0.66         & 84.68$\pm$0.39       & 91.33$\pm$0.30          & 4.0            \\
VGAE             & 83.60$\pm$0.52       & 63.37$\pm$1.21       & 78.23$\pm$1.63         & 87.21$\pm$0.26       & 89.79$\pm$0.09          & 5.2          \\
SSL-GCN         & 57.29$\pm$0.13       & 59.57$\pm$1.77        & 75.06$\pm$0.37         & 80.49$\pm$0.10        & 84.71$\pm$0.95          & 6.8          \\
GraphSage       & 74.30$\pm$1.84      & 60.20$\pm$2.15         & 81.96$\pm$0.74        & 87.05$\pm$0.25         & 89.74$\pm$0.19          & 5.6          \\
GraphMAE        & 85.45$\pm$0.40      & 72.48$\pm$0.77         & 85.74$\pm$0.14        & 88.04$\pm$0.61        & \textbf{93.47$\pm$0.04} & 3.0            \\
S2GAE           & 86.15$\pm$0.25      & 74.60$\pm$0.06         & 86.91$\pm$0.28        & 90.94$\pm$0.08        & 91.70$\pm$0.08          & 2.2          \\
\textbf{HC-GAE} & \textbf{87.97$\pm$0.10} & \textbf{75.29$\pm$0.09} & \textbf{87.56$\pm$0.35} & \textbf{91.07$\pm$0.14} & 92.28$\pm$0.07          & \textbf{1.2} \\ \hline
\end{tabular}
\vspace{-0pt}
\end{table*}

\begin{table*}
\vspace{-0pt}
\centering
 \footnotesize
\caption{Graph classification performance based on accuracy. A.R. is the average rank.}\label{graph_cls_res}
\begin{tabular}{ccccccc}
\hline
Datasets        & IMDB-B                & IMDB-M               & PROTEINS              & COLLAB                & MUTAG                 & A. R.        \\ \hline
WLSK            & 64.48$\pm$0.90~       & 43.38$\pm$0.75       & 71.70$\pm$0.67        & N/A                   & 80.72$\pm$3.00        & 7.75         \\
DGCNN           & 67.45$\pm$0.83        & 46.33$\pm$0.73       & 73.21$\pm$0.34~       & N/A                   & 85.83$\pm$1.66        & 6.25         \\
DiffPool        & 72.6$\pm$3.9          & 47.2$\pm$1.8         & 75.1$\pm$3.5          & 78.9$\pm$2.3          & 85.0$\pm$10.3         & 5.20          \\
Graph2Vec       & 71.10$\pm$0.54        & 50.44$\pm$0.87     & 73.30$\pm$2.05          & N/A                   & 83.15$\pm$9.25        & 5.75         \\
InfoGCL         & 75.10$\pm$0.90        & 51.40$\pm$0.80      & N/A                   & 80.00$\pm$1.30          & 91.20$\pm$1.30          & 3.50          \\
GraphMAE        & 75.52$\pm$0.66        & 51.63$\pm$0.52       & 75.30$\pm$0.39       & 80.32$\pm$0.46          & 88.19$\pm$1.26          & 3.20          \\
S2GAE           & 75.76$\pm$0.62        & 51.79$\pm$0.36       & 76.37$\pm$0.43        & \textbf{81.02$\pm$0.53} & 88.26$\pm$0.76          & 2.00            \\
\textbf{HC-GAE} & \textbf{76.72$\pm$0.60} & \textbf{51.90$\pm$1.47} & \textbf{78.13$\pm$1.37} & 80.41$\pm$0.02          & \textbf{92.38$\pm$1.17} & \textbf{1.20} \\ \hline
\end{tabular}
\vspace{-0pt}
\end{table*}

\begin{figure*}
\vspace{-0pt}
    \centering
    \includegraphics[width=.75\textwidth]{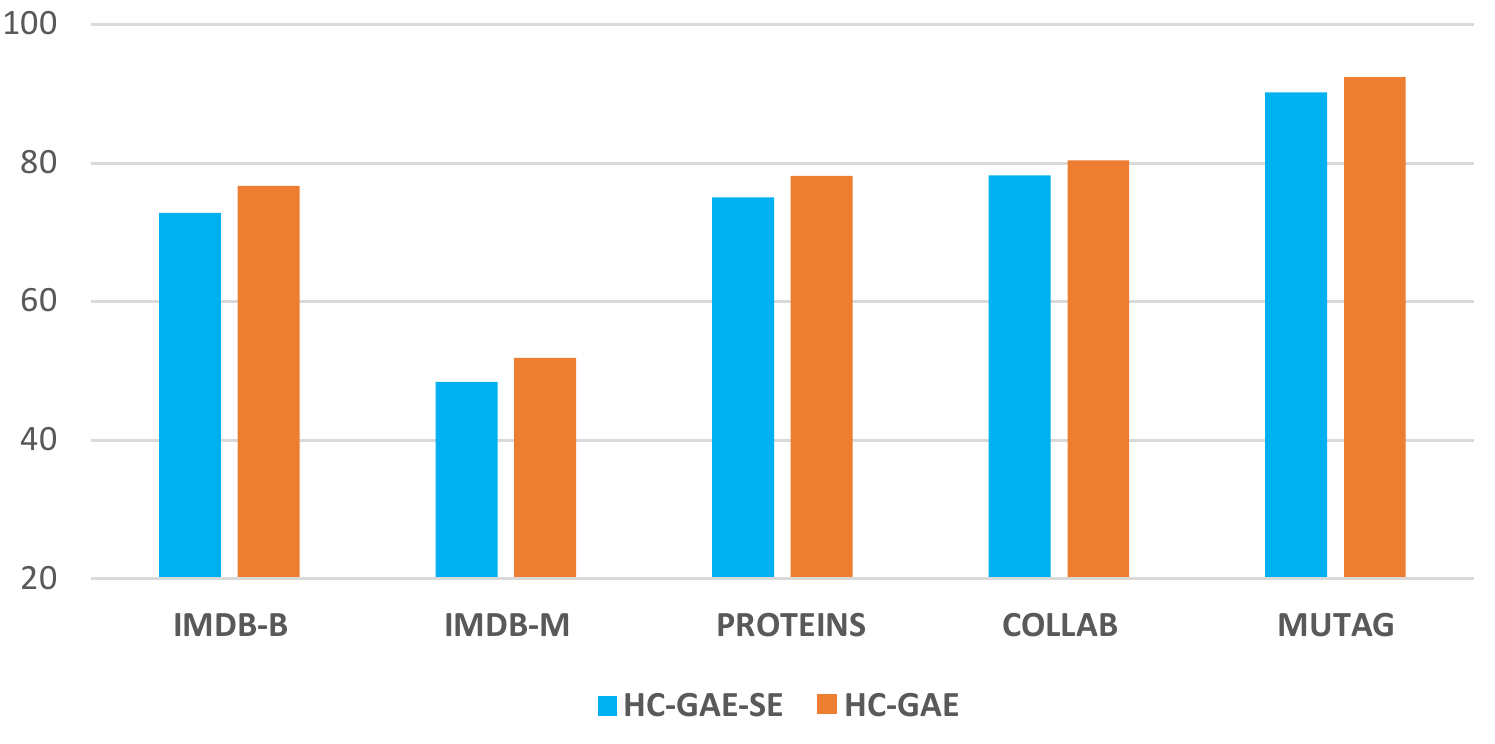}
    \vspace{-10pt}
    \caption{The ablation experiments on graph classification task. }
    \label{fig:ablation}
    \vspace{-0pt}
\end{figure*}
\subsection{Node Classification}
\textbf{Datasets.} For node classification task, we consider 5 real-world datasets (Cora, CiteSeer, PubMed, Amazon-Computers and  Coauthor-CS). To fairly compare our model and the other baselines, we follow the previous study~\cite{node_cls_data} to carry out the related experiments, and utilize the SVM classifier to predict the node labels. We evaluate model performance based on the accuracy score.

\textbf{Baselines.} We compare our model with 6 self-supervised models including DGI~\cite{dgi}, VGAE~\cite{VGAE}, SSL-GCN~\cite{SSL_GCN},  GraphSage~\cite{GraphSAGE}, GraphMAE~\cite{GraphMAE}, S2GAE~\cite{S2GAE}. The reported results of baselines are from previous papers if available.

\textbf{Experimental Setup.} In order to compare methods fairly, we adopt 10-fold cross validation to test all models. And we generally follow the same parameter settings across different baselines. We select the Adam optimizer to optimize the parameters of models in our experiments. And neural network models are trained in 50 epochs. During the training process, we set the hidden dimension of models to 128, the dropout to 0.5.
Specially, for the node classification task, we set the batch size to 1024, the learning rate to $1e-2$. For the graph classification task, we set the batch size to 64, the learning rate to $5e-4$.
For our proposed model HC-GAE, we set the encoder layer $L$ to 3,  the decoder layer $L^{\prime}$ to 3. The node numbers of the three layers in our encoder follow $\{128, 64, 32\}$, and the node numbers of the three layers in our decoder follow $\{32, 64, 128\}$. The experiments were performed on four GeForce RTX 2080 Ti GPUs.

\textbf{Results.} We summarize the results in Table~\ref{node_cls_res}. Obviously, our proposed model HC-GAE can outperform all the baseline models. Only the accuracy of the self-supervised method GraphMAE is a little higher than our model. From these results, we could analyse that the VGAE and its improved methods are effective in node representation learning. The improved GAE methods (e.g. GraphMAE, S2GAE) have a better performance. This verifies that the GAE framework is suitable for node-level representation learning.

\subsection{Graph Classification}
\textbf{Datasets.} For graph classification task, we adopt 5 standard graph datasets (IMDB-B, IMDB-M, PROTEINS, COLLAB, MUTAG). In experiments, we follow the previous study~\cite{GraphMAE}, and feed the resulting graph representations into the SVM classifier for prediction. We also evaluate the performance based on the accuracy score, and report the mean 10-fold cross-validation accuracy with standard deviation.

\textbf{Baselines.} We compare our model with a typical graph kernel model WLSK~\cite{WLSK} based on the subtree invariants, 2 supervised baseline models including DGCNN~\cite{DGCNN} and DiffPool~\cite{DiffPool}, 4 self-supervised baseline models including Graph2Vec~\cite{graph2vec}, InfoGCL~\cite{infogcl}, GraphMAE~\cite{GraphMAE}, S2GAE~\cite{S2GAE}.

\textbf{Results.} We summarize the results in Table~\ref{graph_cls_res}. Similar to the results on the node classification task, our performance on the graph classification is outstanding in the experiments. Only on the COLLAB dataset, S2GAE slightly outperforms our model. With these comparisons of the node classification and the graph classification, we could analyse the reasons why the effectiveness of our model is obvious.
The GAEs such as GraphMAE focusing on the graph feature reconstruction could obtain outstanding performance on the specific dataset of the node classification. However, its performance on the graph classification is not as good as the other GAEs (e.g., S2GAE, HC-GAE) focusing on graph features and structure. This verifies that the topological missing disturbs GraphMAE while the combination of the assignment strategies and the re-design loss improves our proposed HC-GAE for multiple downstream tasks.
In the hierarchical methods such as DGCNN and DiffPool, there is an assignment process when the input graphs are compressed into the coarsened graphs. However, DGCNN adopts the top-k strategy to assign nodes, and  DiffPool utilizes the hard assignment. Since these methods \textbf{cannot} prevent the information passing causing the over-smoothing, their performance is limited in the experiments. However, the separated subgraphs proposed in our encoder avoid the information passing, realizing the improvement of the GAE.

\subsection{Ablation Study}
In order to analyse the effectiveness of our encoder, we replace the separated subgraphs and the hard assignment strategy in the encoding with the soft assignment strategy. The comparison results are shown in the Figure~\ref{fig:ablation}. In this experiment, we define our model with the soft assignment in the encoder as \textbf{HC-GAE-SE}. We observe that the performance of HC-GAE-SE on graph classification is lower than ours. Compared to the HC-GAE-SE, our vanilla model have two factors which strengthen the training. First, the separated  subgraphs prevents the encoding from the over-smoothing causing the fall of the performance. The subgraph generation combining with the hard assignment makes the message pass within the subgraph. Secondly, the loss value calculation relying on the local graph information. Without the original encoder, $\mathcal{L}_{\mathrm{HC-GAE}}$ missing the local loss $\mathcal{L}_{\mathrm{local}}$ cannot allow the encoding process to extract the local information.

\section{Conclusion}
\label{conclusion}

In this paper, we have proposed a novel HC-GAE model to effectively learn multi-level graph representations for various downstream tasks, i.e., the node classification and the graph classification. During the encoding process, we have adopted the hard node assignment to decompose a sample graph into a family of separated subgraphs, that can be compressed into the coarsened nodes for the resulting coarsened graph. On the other hand, during the decoding process, we have utilized the soft node assignment to reconstruct the original graph structure. During the encoding and decoding processes, the proposed HC-GAE can effectively extract features hierarchical graph representations. The re-designed loss function has balanced the training of the encoder and the decoder. In the experiments, we have evaluated the performance of the proposed HC-GAE model for either node classification or graph classification. The experimental results have demonstrated the effectiveness of the proposed model.




\balance


\bibliographystyle{IEEEtran}
\bibliography{mybibfile}

\end{document}